# A Deep Learning-Based Approach to Extracting Periosteal and Endosteal Contours of Proximal Femur in Quantitative CT Images

Yu Deng[1], Ling Wang[2], Chen Zhao[3], Shaojie Tang[1], Xiaoguang Cheng[2], Hong-Wen Deng[4], Weihua Zhou[3]

[1]School of Automation, Xi'an University of Posts and Telecommunications, Xi'an, Shaanxi 710121, China
[2]Department of Radiology, Beijing Jishuitan Hospital, Beijing 100035, China
[3]College of computing, Michigan Technological university, Houghton, MI 49931, USA
[4]Tulane University, Department of Biomedical Engineering, New Orleans, LA 70118 USA

Yu Deng and Ling Wang have contributed equally.

**Corresponding authors:**
Shaojie Tang, PhD
School of Automation, Xi'an University of Posts and Telecommunications, Xi'an, Shaanxi 710121, China
E-Mail: tangshaojie@xupt.edu.cn
X. Cheng, PhD
Department of Radiology, Beijing Jishuitan Hospital, Beijing 100035, China.
E-Mail: xiao65@263.net

Objective: Automatic CT segmentation of proximal femur is crucial for the diagnosis and risk stratification of orthopedic diseases. In this study, we proposed an approach based on deep learning for the automatic extraction of the periosteal and endosteal contours of proximal femur in order to differentiate cortical and trabecular bone compartments.

Methods: A three-dimensional (3D) end-to-end fully convolutional neural network, which can better combine the information between neighbor slices and get more accurate segmentation results, was developed for our segmentation task. 100 subjects aged from 50 to 87 years with 24,399 slices of proximal femur CT images were enrolled in this study. The separation of cortical and trabecular bone derived from the QCT software MIAF-Femur was used as the segmentation reference. We randomly divided the whole dataset into a training set with 85 subjects for 10-fold cross-validation and a test set with 15 subjects for evaluating the performance of models.

Results: Two models with the same network structures were trained and they achieved a dice similarity coefficient (DSC) of 97.87% and 96.49% for the periosteal and endosteal contours, respectively. To verify the excellent performance of our model for femoral segmentation, we measured the volume of different parts of the femur and compared it with the ground truth and the relative errors between predicted result and ground truth are all less than 5%. It demonstrated a strong potential for clinical use, including the hip fracture risk prediction and finite element analysis.

Significance: This approach is helpful to measure the BMD of cortical bone and trabecular bone respectively, and to evaluate the bone strength based on finite element analysis.

1. Introduction

Hip fracture is the most serious fragility fracture, associated with a high mortality and a significant decline in mobility and quality of life in the elderly [1]. Using quantitative computed tomography (QCT) to measure bone mineral density (BMD) and geometric parameters to predict fracture risk and to diagnose osteoporosis is commonly employed in research studies or clinical trials [2, 3]. Based on the three-dimensional (3D) imaging technology of QCT, trabecular and cortical bones can be separated, and geometrical parameters such as cortical thickness can be measured. It is essential to understand well the role of trabecular and cortical compartments in hip fracture risk prediction [4, 5]. Meanwhile, bone strength in vivo could be accurately assessed by finite element analysis (FEA) based on QCT [6], which has the potential to improve the hip fracture risk assessment [7, 8]. Further, accurate femur segmentation of the periosteal and endosteal contours directly affects the subsequent analysis and processing with computer-aided diagnosis (CAD) systems.

For the accurate physical BMD measurements and bone strength assessments, the accurate bone segmentation of femur is a key prerequisite. However, the CT segmentation methods commonly used for femur are either time-consuming or lack of accuracy [9]. The segmentation methods of femoral CT include manual segmentation, computer interactive segmentation and automatic segmentation. Manual segmentation and computer interactive segmentation have high requirements on expertise and experience, and there are inevitable operator errors. Moreover, the researches on femur CT segmentation mainly focus on the periosteal contours of femur. Therefore, the automatic segmentation of femur CT is desired for clinical practice.

The automatic segmentation methods commonly used for femur mainly include watershed methods [10], shape-model-based methods [11], and level set method [12]. However, these methods are limited by segmentation accuracy and speed, so they are unlikely to be effectively applied in a clinical practice. With the enhancement of the computing ability of modern computers and the successful application of artificial intelligence technology, deep learning is popularly used to process various sorts of images [13, 14]. Zhu et al. divided the femur into three parts according to their spatial positions, and segmented the femur using conditional generative adversarial network (CGAN) [15]. Chen et al. proposed a 3D feature enhancement network for femur segmentation including edge detection task and multi-scale feature fusion [16].

The existing deep learning methods for femur segmentation mainly focus on the segmentation of periosteal contour, which motivates us to develop a fast and reliable segmentation approach for segmenting both periosteal and endosteal contours. In this paper, we proposed a deep convolution neural network for segmenting both periosteal and endosteal contours of proximal femur automatically and accurately.

## 2. Materials and Methods

### 2.1 Dataset

Our study involved 100 subjects aged from 50 to 87 years with 24,399 slices of proximal femur CT images obtained from Jesusita Hospital, Beijing, China, between August 2018 and June 2019. The CT scanning parameters were 120 kV, 250 mAs and 1 mm slice thickness. The ground truth used in the training was manually obtained from Jishuitan Hospital through a semi-automatic interactive segmentation software tool of the Medical Image Analysis Framework Option Femur (MIAF Femur v7.1.0MRH, Klaus Engelke, Erlangen) [17]. During the experiments for the entire cohort, 85% of the subjects were used in 10-fold cross-validation for training and internal validation, and to select the optimal parameters of the proposed models; the rest of the subjects were used to evaluate the performance of models.

### 2.2 Image preprocessing and data augmentation

Since the original images were only annotated with the left femurs, we manually cropped the sizes of these images from the original 512 × 512 to 192 × 192, so that the cropped images only contain the left femurs. This not only reduces the requirements on the memory size in GPU during the training process, but also eliminates more unnecessary interference factors. One example of cropping a femoral CT image is shown in Fig. 1.

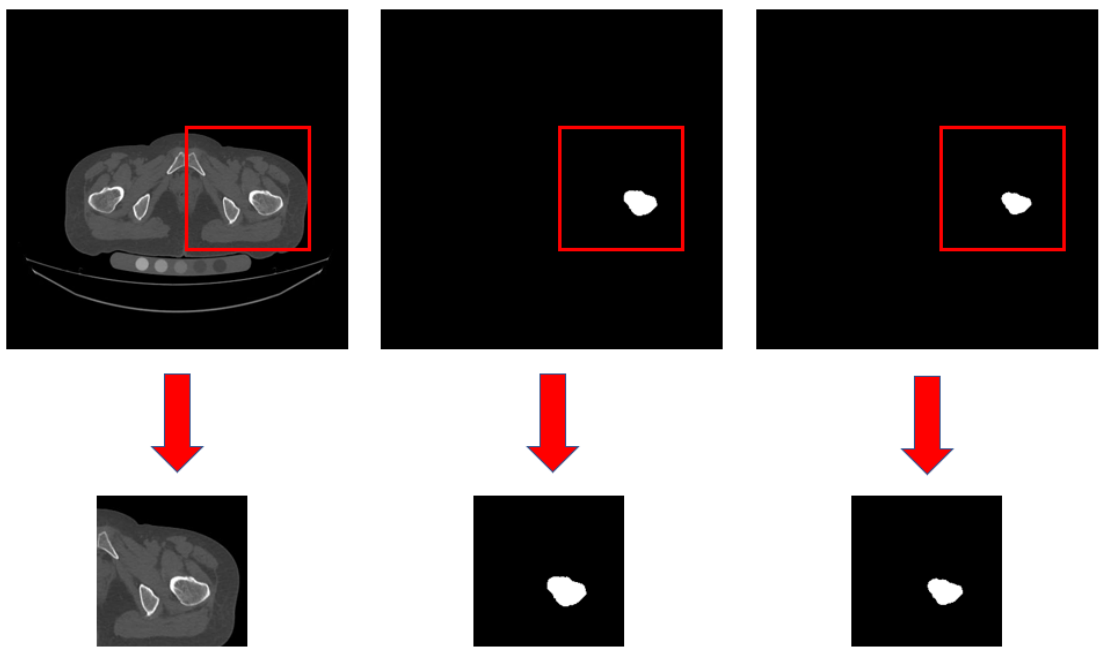

Fig.1 Cropping process of femoral CT slice. From left to right, the three columns correspond to the original femoral slice, the ground truth of periosteal contour, and the ground truth of endosteal contour, respectively. Note that the endosteal contour is enclosed by the periosteal contour.

Then, we performed data augmentation on the training data, by adding Gaussian noise, performing a brightness transformation, and carrying out a 3D geometrical scaling, to increase the capacity of the training samples, thereby improving the generalization ability and robustness of the model to be trained [18].

### 2.3 Proposed deep learning network

Due to the superior performance of 3D neural networks in medical image segmentation, we carried out a 3D end-to-end neural network based on the well-known V-net [19]. 3D neural networks can effectively exploit the contextual information among neighbor slices of medical images, which can ensure the continuity of the change among the image masks in neighbor layers [20]. The V-net for image segmentation can be abstracted as an encoder-decoder structure. The encoder is used to extract features and reduce the size of feature maps, while the decoder is used to decode the features and restore the size of feature maps. The detailed structure of the 3D V-net for proximal femur segmentation is shown in Fig. 2.

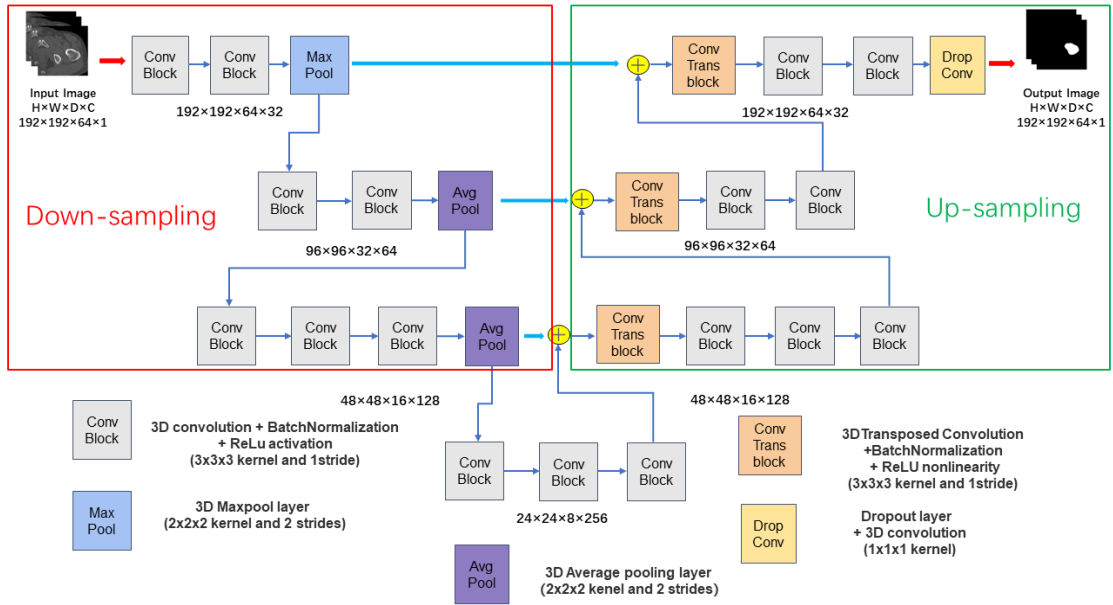

Fig. 2. Detailed structure of the 3D V-net for proximal femur segmentation.

The V-net used is essentially of a symmetric architecture, wherein the left part of the network consists of a down-sampling path, while the right part is an up-sampling path to restore the volume to the input size. We set the kernel size of convolution layers to $3 \times 3 \times 3$, which reduces the total amount of the trained parameters to ~8.6M, and greatly decreases the training time. Each convolution block contains a batch normalization (BN) and ReLU layer to prevent overfitting and enhance the nonlinearity of the neural network. The output of the neural network is a probability map of the same size as in the input volume. Finally, the Otsu algorithm was used to transform the probability map into a binary segmentation result.

Our model training was performed in the environment of Python 3.6, tensorflow 1.13.1 with an Nvidia Geforce GTX 1080Ti GPU. The ADAM optimizer was used for the training process [21]. The initial learning rate is 0.0001, the epoch is 800, the patch size is 192 × 192 × 64 and the batch size is 1. In order to make our model better predict 3D volume data with arbitrary slices, we used a sliding window technique to get the final prediction results. In other words, if a subject contains S slices, our model will run S-64+1 times to reach the final prediction. We used a softmax cross entropy loss function for training as follows,

$$L_{ce}(y,p) = -\frac{1}{N}\sum_{n=1}^{N}\sum_{c=1}^{C} y_{n,c} \log(p_{n,c}) \tag{1}$$

where $y_{n,c} \in [0,1]$ is the label of ground truth, $p_{n,c} \in [0,1]$ is the estimated probability, $N$ denotes the number of voxels, and $C$ represents the number of object classes.

## 2.4  Evaluation metrics

We used the data mentioned in the subsection 2.1 for the contour segmentation of proximal femur. In order to improve the generalization ability of our model and overcome the overfitting problem, a 10-fold cross-validation was performed and the model with the highest dice similarity coefficient (DSC) in the cross-validation was selected as our final trained model. DSC is defined as follows,

$$DSC = \frac{2TP}{FP + 2TP + FN} \tag{2}$$

where TP, FP and FN are the detected numbers of true positives, false positives and false negatives, respectively. We defined voxels with the femur and background voxels as positive and negative results, respectively. DSC is essentially a similarity measure between two sets, which is usually used to calculate the similarity between the prediction result and ground truth.

To quantify the performance of the trained models, the DSC mean, average surface distance (ASD), sensitivity and specificity of the segmentation results were adopted [18]. Sensitivity, also called true positive rate (TPR) and recall, measures the portion of proximal femur voxels in the ground truth that are also identified as proximal femur voxels by the automatic segmentation being evaluated. Sensitivity is defined as,

$$sensitivity = \frac{TP}{TP + FN} \tag{3}$$

Similarly, specificity measures the portion of background voxels in the ground truth that are also identified as a background voxel by the automatic segmentation. Specificity is defined as,

$$specificity = \frac{TN}{TN + FP} \tag{4}$$

ASD is a common measure for the evaluation of medical image segmentation results. By defining the shortest Euclidean distance of an arbitrary voxel $v$ to a surface $S$ by $d(v,s) = \min_{s \in S}\|v - s\|$, ASD can be written as,

$$\text{ASD} = \frac{1}{N_S + N_G}\left(\sum_{x_S \in S} d(x_S, G) + \sum_{x_G \in G} d(x_G, S)\right) \tag{5}$$

where $N_S$ and $N_G$ are the voxel numbers on the surfaces of the automatic segmentation and ground truth, respectively.

In addition, we calculated the volumes of femur and cortical bone in the femoral segmentation and compared the volume of each compartment measured from our segmentation with that from the corresponding ground truth in all the 15 test subjects. Mean absolute error (MAE), root mean square error (RMSE) and relative error (RE) indexes were reported as the evaluation metrics. They are defined as follows,

$$\text{MAE} = \frac{1}{m}\sum_{i=1}^{m}|y_i' - y_i| \tag{6}$$

$$\text{RMSE} = \sqrt{\frac{1}{m}\sum_{i=1}^{m}(y_i' - y_i)^2} \tag{7}$$

$$\text{RE} = \frac{|y_i' - y_i|}{y_i'} \times 100\% \tag{8}$$

where $y_i'$ and $y_i$ denote the volumes calculated from the ground truth and our prediction result, respectively.

## 3. Experimental results

### 3.1 Periosteal contour segmentation

Training the model for segmenting the periosteal contour of femur converged after 400 epochs. The average DSC of the segmentation results for eight subjects in the validation set is 97.87%. As illustrated in Fig. 3, the periosteal contours from our model match well with those from the ground truth.

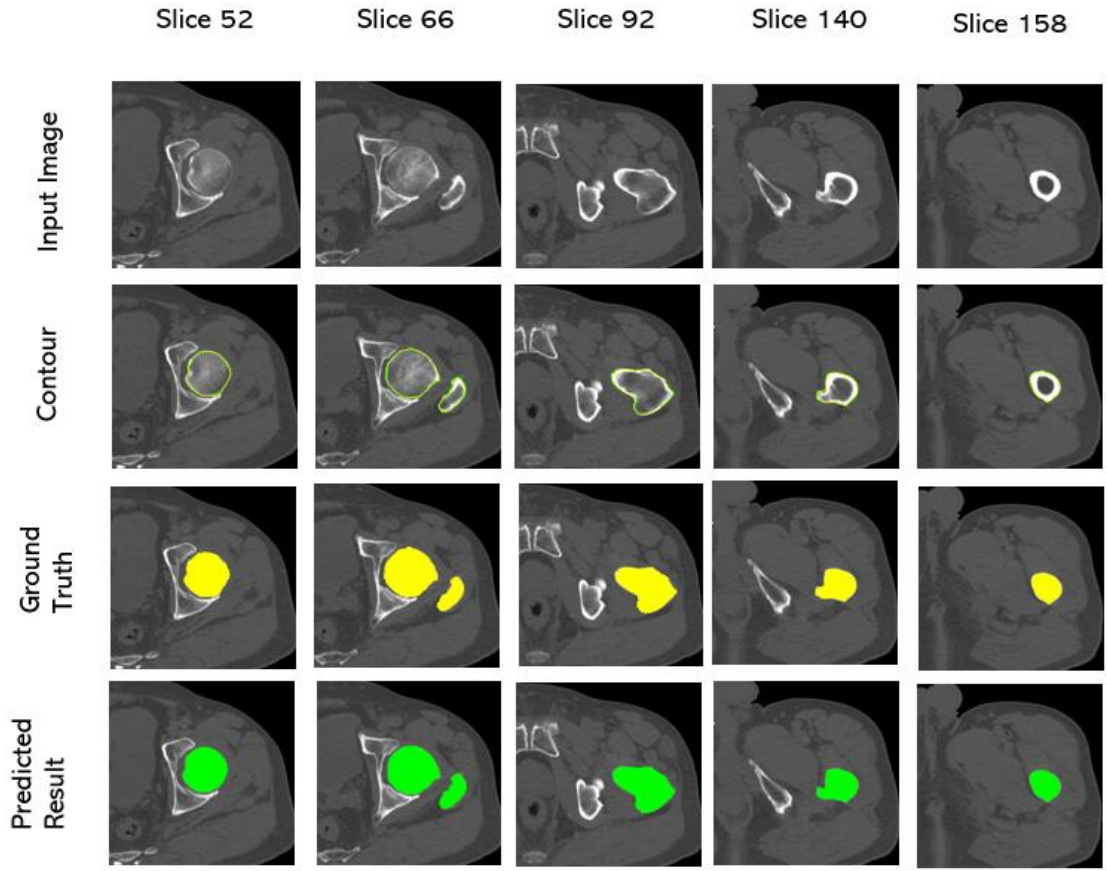

Fig. 3. Periosteal contour segmentation. The green and yellow contours represent the results segmented with the 3D V-net and ground truth, respectively.

### 3.2  Endosteal contour segmentation

Since the endosteal region is enclosed by the periosteal contour (as illustrated in Fig. 1), the region inside the periosteal contour from Section 3.1 is used as the input of endosteal contour segmentation. A method similar to that described in the Section 3.1 was then adopted to train the model of endosteal contour segmentation. The training procedure converged within 240 epochs. The average DSC of the segmentation results for eight subjects in the validation set is 96.49%. As illustrated in Fig. 4, the endosteal contours from our model match well with those from the ground truth.

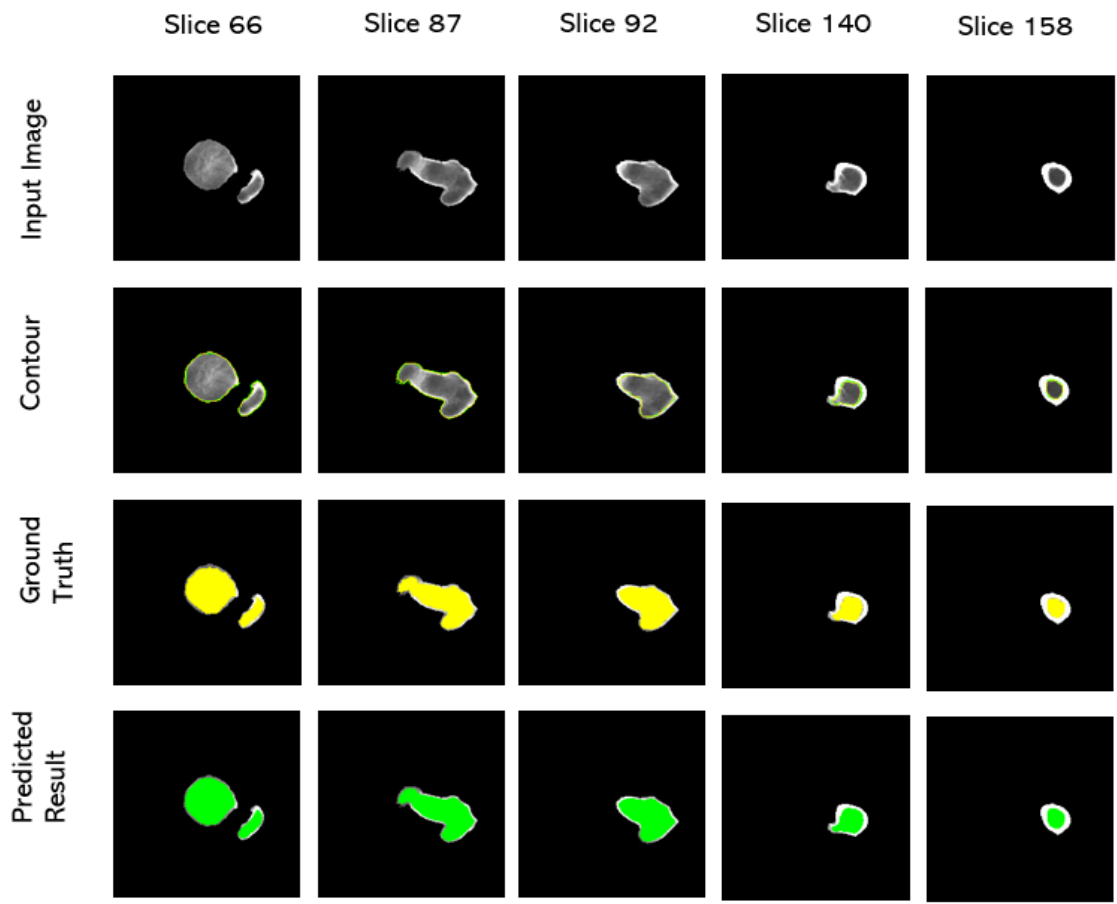

Fig. 4. Endosteal contour segmentation. The green and yellow contours represent the results segmented with the 3D V-net and ground truth, respectively.

**3.3  3D visualization**

From the above results, we can see that the trained models can well segment the periosteal and endosteal contours of proximal femur. In order to better demonstrate the segmentation results, two subjects are visualized in 3D as shown in Figs 5.

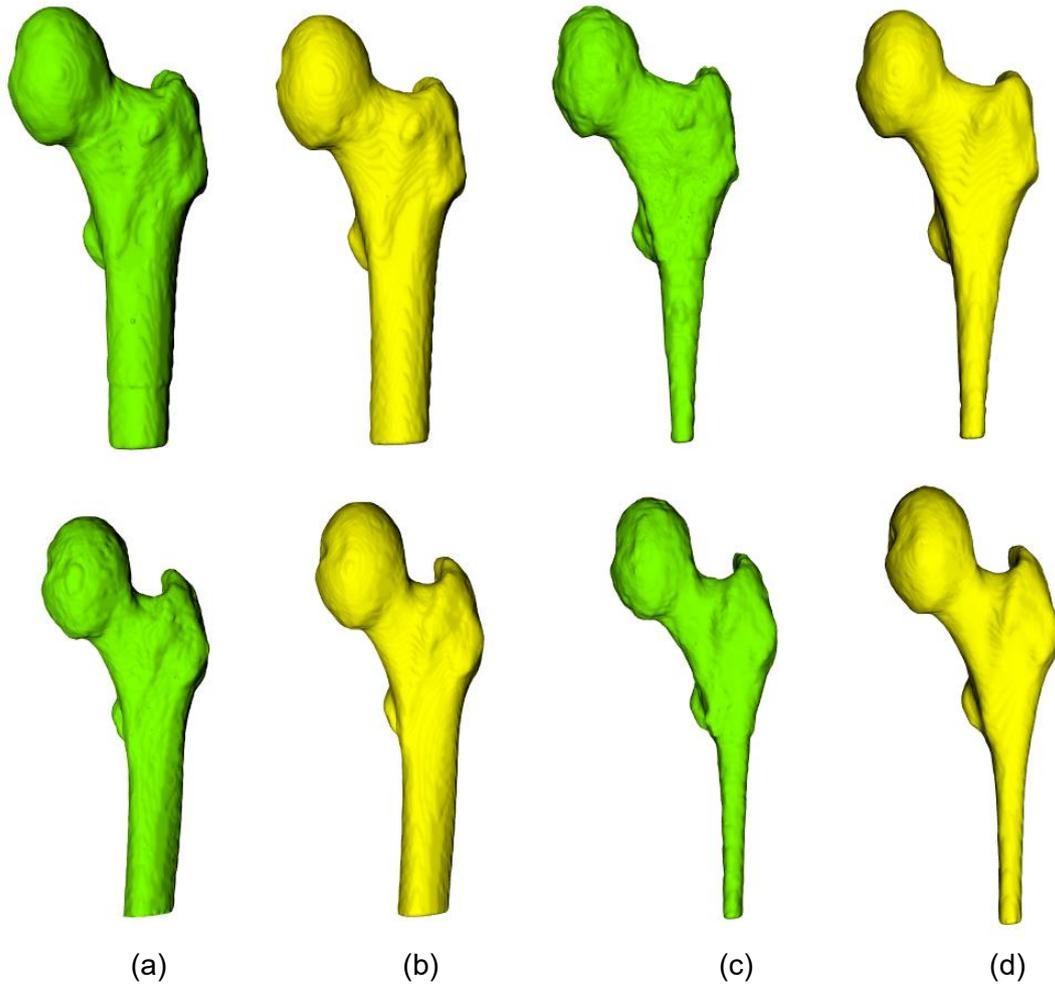

(a) (b) (c) (d)

Fig. 5. 3D visualization of the periosteal and endosteal surfaces of proximal femurs. (a) The periosteal surfaces from the ground truth; (b) the periosteal surfaces from automatic segmentation; (c) the endosteal surfaces from the ground truth; and (d) the endosteal surfaces from automatic segmentation. Note that the two rows correspond to two subjects, respectively.

For each surface voxel in the segmentation results, the Euclidean distance from the nearest contour voxel in the ground truth was calculated as surface distance error. In Fig. 6, the distance errors of two subjects are visualized with color.

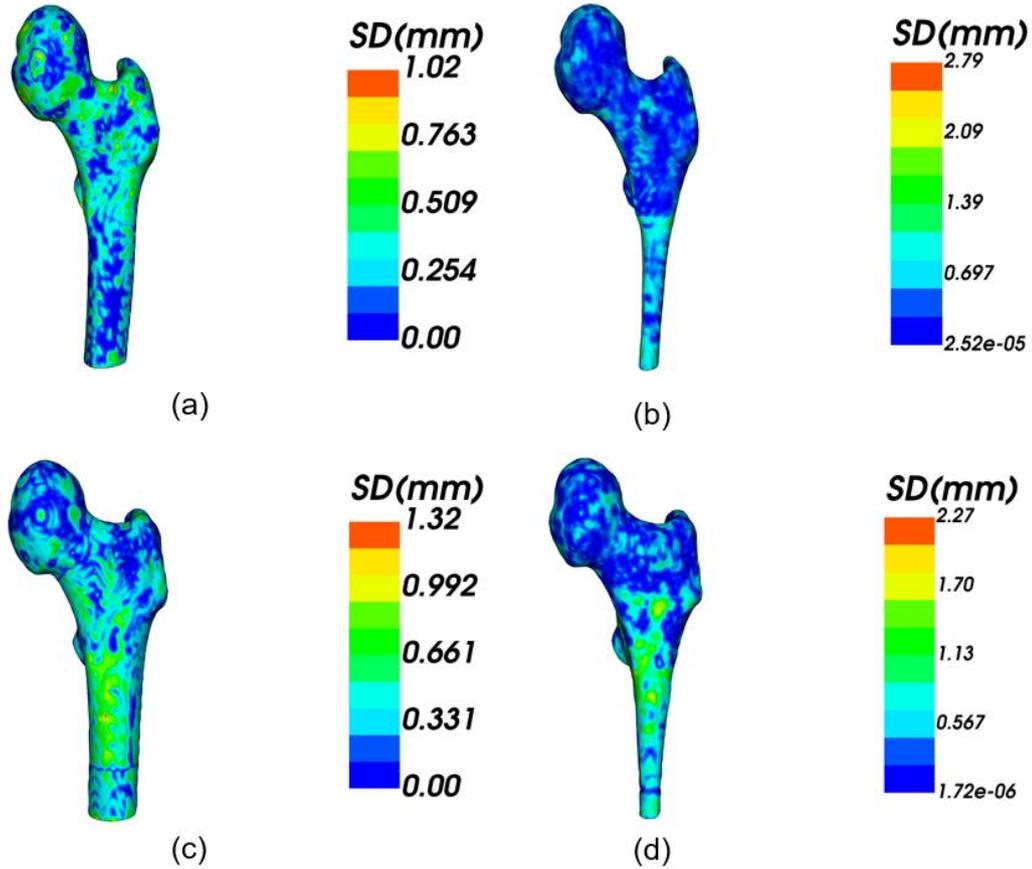

Fig.6. Visualization of surface distance errors. (a) and (c) the distance errors of the periosteal surfaces between automatic segmentation and the ground truth; (b), (d) the distance errors of the endosteal surfaces between automatic segmentation and the ground truth. Note that the two rows correspond to two subjects, respectively.

4. **Performance evaluation**

Table 1 shows the DSC, ASD, SP and SN of the periosteal and endosteal contours in the test dataset. Overall, there is an excellent agreement in segmentation between our method and the ground truth. Both DSCs are as high as >0.96. The ASDs are small than 0.32 mm, which are small compared to the original voxel size of $0.8 \times 0.8 \times 0.987$ cm$^3$.

Table 1. Quantitative results evaluated with the test set.

| Evaluation parameters | DSC | ASD (mm) | SP | SN |
|---|---|---|---|---|
| Periosteal surface segmentation | 0.9787 | 0.2011 | 0.9887 | 0.9939 |
| Endosteal surface segmentation | 0.9649 | 0.3117 | 0.9984 | 0.9838 |

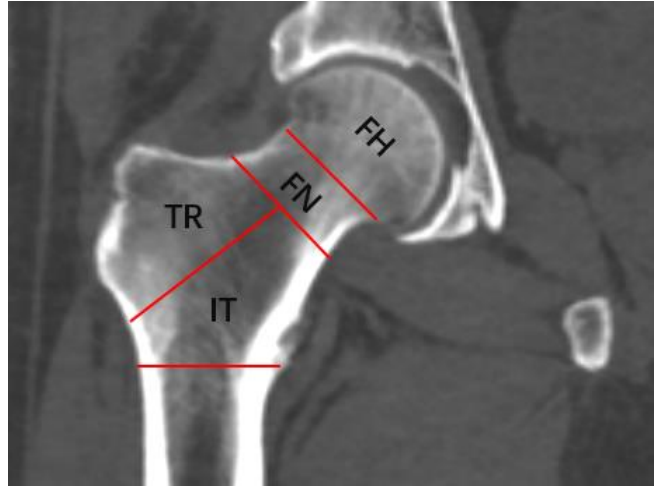

Fig. 7. Regions of a femur. FH, femur head; FN, femur neck; TR, trochanter; IT, intertrochanter.

Further, the femur was divided into four parts defined by the Mindways software (Mindways Software Inc, Austin, TX) [1], including femur head (FH), femur neck (FN), trochanter (TR), intertrochanter (IT), as shown in Fig. 7. The volumes of two regions, neck and the combination of neck, TR and IT were measured. The volumes of femoral head, femoral neck and the combination, were also evaluated. The quantitative comparison between our prediction results and the ground truth for each part is listed in Table 2. Overall, the mismatch between our prediction and the ground is relatively small.

Table 2. Quantitative comparison between our prediction results and the ground truth. Note that "Combination" refers to the combination of neck, trochanter and intertrochanter. MAE, mean absolute error; RMSE, root mean square error; RE, relative error.

| Metric<br>Target | Ground truth (cm$^3$) | MAE (cm$^3$) | RMSE (cm$^3$) | RE(%)<br>(mean, min, max) |
|---|---|---|---|---|
| Femur head | 38.91±17.09 | 1.1326 | 1.2790 | (3.09, 0.09, 6.85) |
| Femur neck | 6.83±5.35 | 0.2443 | 0.2701 | (3.69, 0.93, 5.75) |
| Combination | 71.17±20.42 | 2.3452 | 2.5422 | (3.30, 0.13, 5.04) |
| Cortical bone in femur neck | 1.62±0.97 | 0.0771 | 0.0880 | (4.65, 1.07, 9.45) |
| Cortical bone in combination | 16.84±3.85 | 0.6146 | 0.7868 | (4.13, 0.28, 11.19) |

In Fig. 8, the relative errors in five regions of 15 test subjects are visualized to further demonstrate the performance of our segmentation. The horizontal axis indicates the volume of the proximal femur in the ground truth while the vertical axis represents the relative error of the proximal femur volume measured from our prediction as compared to the ground truth. The 15 points in the figure correspond to 15 test subjects, respectively.

According to Table 2 and Fig. 8, the average relative errors of all the five femoral parts are less than 5%; the biggest error occurs in the cortical bone of femur neck and the smallest in the femoral head. Overall, our segmentation model is accurate for most subjects.

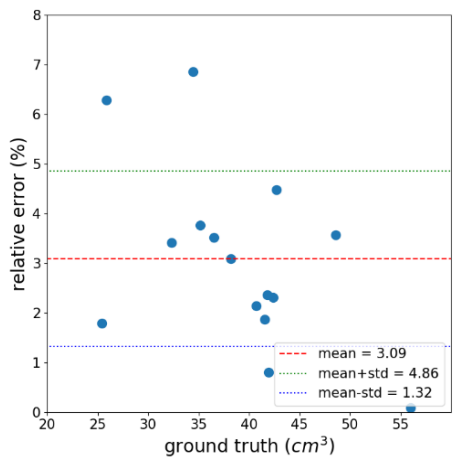

(a)

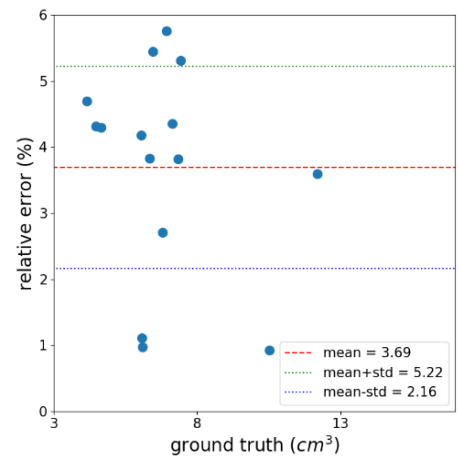

(b)

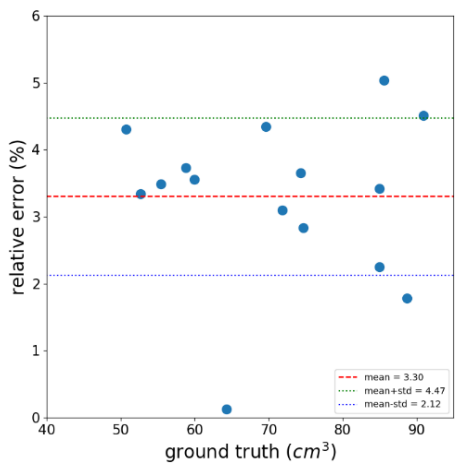

(c)

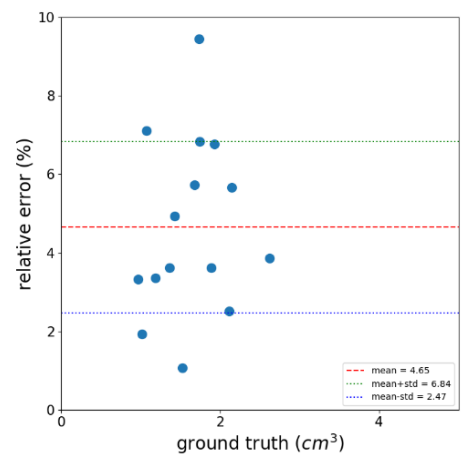

(d)

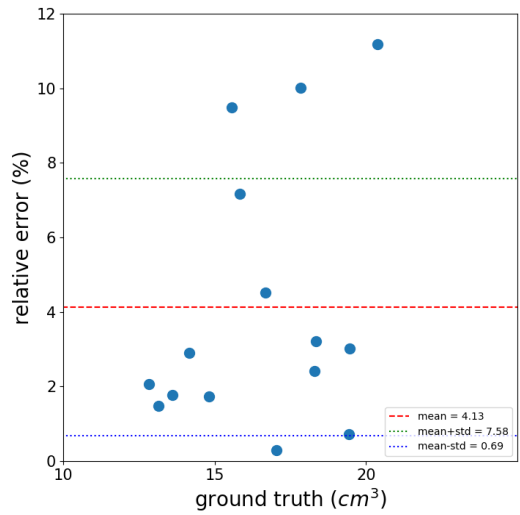

(e)

Fig. 8. Relative errors between the volumes of segmentation results and the ground truth. (a) The femur head, (b) femur neck, (c) combination, (d) cortical bone in femoral neck, and (e) cortical bone in the combination.

## 5. Discussions

Automatic femur segmentation in CT for extracting periosteal and endosteal surfaces is challenging task. The main difficulties to achieve an accurate and automatic segmentation for femur include the weak femoral border, narrow inter bone spacing, low quality CT scanning and leg posture of subjects. Fig. 9 shows two examples, including a weak femoral boundary between hip and femur (Fig. 9a), and the narrowness of the space in a joint (Fig. 9b). The aim of the present study is to develop and validate a fully automatic femur segmentation method that can address this challenge such that it can be used in different computer-assisted orthopedic disease diagnosis. The method was verified successfully with 15 test subjects, proving to be effective in performance.

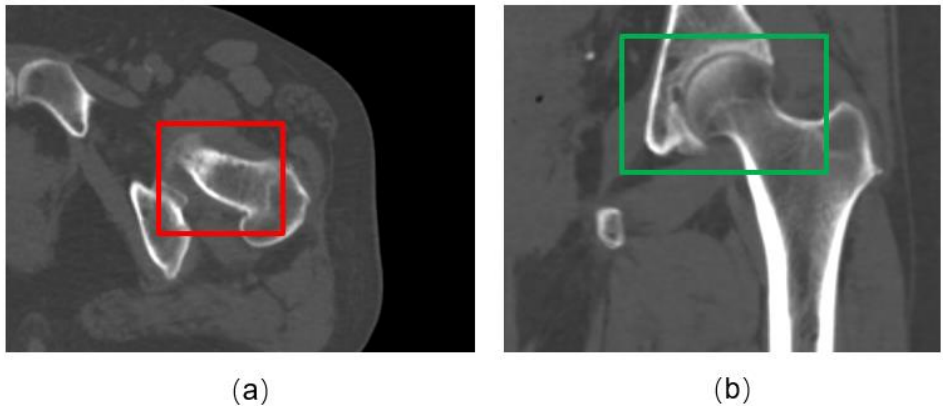

Fig 9. Examples of challenging femur segmentation. (a) Blurry bone boundaries. (b) Narrow inter-bone spacing.

In our method, a deep learning network was firstly implemented to segment the periosteal femur contour. Then, the region enclosed by the periosteal contour was used as the input of endosteal contour segmentation. Compared with the cropped CT image as the input image, the DSC of endosteal contour segmentation was improved from 92.56% to 96.49%.

Since a 3D neural network is used and the computation is very consumptive, it is unrealistic to use images with an in-plane size of 512 × 512 as training samples. Therefore, we cropped the original images into a size of 192 × 192, which reduced the computational burden when training the model, eliminated much unnecessary feature information and mitigated the training difficulty. Although the ground truth we used were generated by experienced radiologists with the help of a sophisticated software tool and the femoral contours were well delinearted, there are still limitations and discrepancies in the ground truth between operators. As shown in Fig. 5, the annotated contours in some regions in the ground truth are not sufficiently continuous and smooth. However, our 3D segmentation model can effectively exploit the continuity of the neighbor image slices in the 3D space, so the femoral surfaces in the final segmentation results are enforced to be smooth as expected.

Historically, there were various researches in segmenting and analyzing femur. Chen et al. used a 3D feature enhancement network to segment the periosteal contour of femur and validated their model in 30 subjects. Their method achieved a DSC of 0.9688, and it took 0.93 s to segment a CT volume[16]. As a comparison, our proposed approach achieved a DSC of 0.9787, and only 0.31 s was required to segment a CT volume. In addition, Zhu et al proposed a multi-stage method to segment the femoral head and a DSC of more than 0.98 was achieved [15]. In their method, one CNN is firstly used to classify all the CT slices into upper, middle or lower parts and then three conditional generative adversial networks are needed for segmentation of each part. This complicated multi-stage srategy brings difficulties to parameter tuning and increases both the training and test time. Furthermore, we compared our approach with the traditional image processing methods. The method based on adaptive shape matching (ASM) proposed by Almeida [11] segmented the femur with medullary canal; the DSC was 0.94 ± 0.016, and the average distance error was 1.014 ± 0.474 mm. Of a particular note, this method is prone to local unexpected variances of the femoral shape due to the nature of ASM segmentation. The qualitative and quantitative evaluation shows that our method has excellent performance in automatic segmentation of proximal femur in QCT images. It is worth noting that due to the fact that different datasets were used in different segmentation methods, a direct comparison of the performance among different methods is difficult. Therefore, as a comparison, the U-Net network [23], which is widely used in medical image segmentation, was implemented as the benchmark model in our study. The U-Net segmentation models achieved the DSCs of 0.9653 and 0.9376 for periosteal and endosteal contour, respectively. Accordingly, our proposed apporach has an improved performance.

Although our approach achieved a high accuracy in 15 test subjects, the present study has still several limitations. First, in some subjects, the error in the cortical neck is much bigger than that in any other part. The endosteal boundary, particularly in the femoral neck, may often be unclear. As shown in Fig 10, the input slices were acquired in the cross-sectional direction (Fig 10c); however, a clear endosteal boundary of femoral neck would be acuqired along the femoral neck axis (Figure 10b). As a result, both the segmentation and related measurement were influenced. Second, only 15 subjects were used to validate the effectiveness of our method. Although this proof-of-concept study demonstrated the potential of our proposed method, the test set with a relatively limited size can not fully represent the variation of a population. Therefore, in order to be applied in clinical practice, a larger CT data set is desired to further train and validate our proposed method.

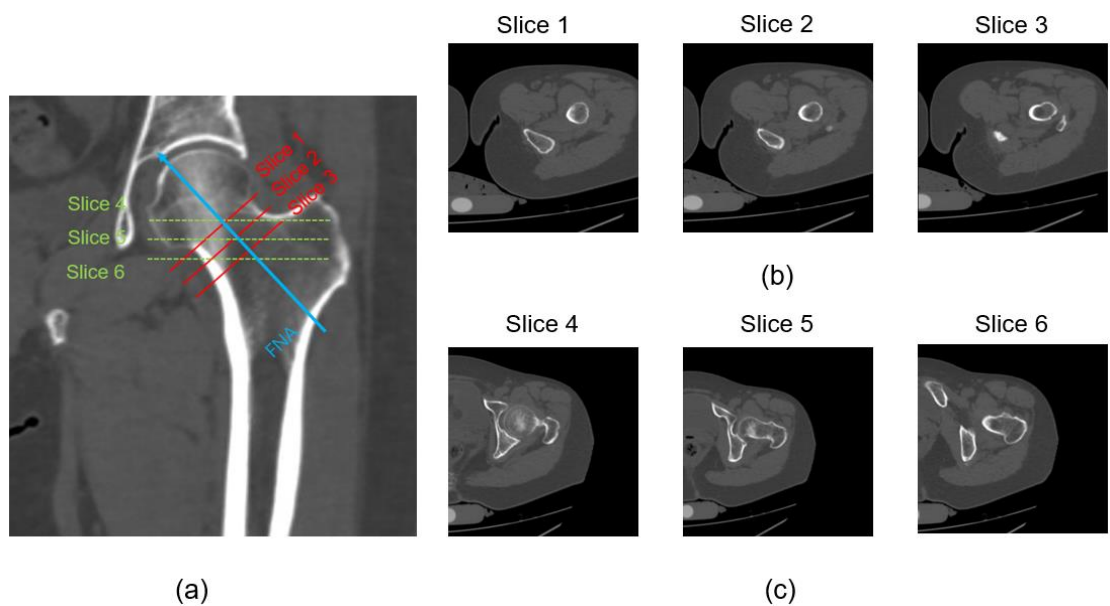

Fig 10. Femoral neck slices. (a) Location of femoral neck slices; (b) Slices along the FNA; and (c) slices on transverse planes. FNA, femoral neck axis.

In future, we will add more subjects with fractures as the training data to improve the generalizability of the trained model, so that it can accurately segment the medical images of proximal femurs with fractures. If this method can be further validated on a dataset with sufficient proximal femurs, it will be expected helpful to specific clinical applications, such as BMD measurement for cortical and trabecular bone, and bone strength assessment based on FEA [24].

## 6. Conclusions

We presented a deep neural network based on 3D V-net for automatic segmentation of proximal femur in CT images. It can segment both the periosteal and endosteal contours effciently and effectively. Our segmentation models achieved a DSC of 0.9787 and 0.9649 for periosteal and endosteal contours, respectively. Using the two models we trained, it only took around 0.3 s to segment a CT volume. Furthermore, the entire and regional volumes of femur measured from our method agreed well with the ground truth. Our method has a great potential for clinical use.

**Acknowledgement**

This work was supported by Shaanxi Provincial Natural Science Foundation of China (Grant No. 2020SF-377), and by Xi'an Key Laboratory of Advanced Controlling and Intelligent Processing(ACIP), China (2019220714SYS022CG044).

Ling Wang's work was also supported the National Natural Science Foundation of China (Grant Nos. 81901718, 81771831, 81971617), the Beijing Natural Science Foundation-Haidian Primitive Innovation Joint Fund (Grant No. L172019), Beijing JST Research Funding (Grant No. 8002-903-02).

**References**

1. Johannesdottir, F., Turmezei, T., & Poole, K. E. (2014). Cortical bone assessed with clinical computed tomography at the proximal femur. Journal of Bone and Mineral Research, 29(4), 771-783.

2. Engelke, K., Lang, T., Khosla, S., Qin, L., Zysset, P., Leslie, W. D., ... & Schousboe, J. T. (2015). Clinical use of quantitative computed tomography (QCT) of the hip in the management of osteoporosis in adults: the 2015 ISCD official positions—part I. Journal of clinical densitometry, 18(3), 338-358.

3. Engelke, K. (2017). Quantitative computed tomography—current status and new developments. Journal of Clinical Densitometry, 20(3), 309-321.

4. Black, D. M., Bouxsein, M. L., Marshall, L. M., Cummings, S. R., Lang, T. F., Cauley, J. A., ... & Orwoll, E. S. (2008). Proximal femoral structure and the prediction of hip fracture in men: a large prospective study using QCT. Journal of Bone and Mineral Research, 23(8), 1326-1333.

5. Manske, S. L., Liu-Ambrose, T., Cooper, D. M. L., Kontulainen, S., Guy, P., Forster, B. B., & McKay, H. A. (2009). Cortical and trabecular bone in the femoral neck both contribute to proximal femur failure load prediction. Osteoporosis international, 20(3), 445-453.

6. Pahr, D. H., & Zysset, P. K. (2016). Finite element-based mechanical assessment of bone quality on the basis of in vivo images. Current osteoporosis reports, 14(6), 374-385.

7. Gong, H., Zhang, M., Fan, Y., Kwok, W. L., & Leung, P. C. (2012). Relationships between femoral strength evaluated by nonlinear finite element analysis and BMD, material distribution and geometric morphology. Annals of biomedical engineering, 40(7), 1575-1585..

8. Bisheh, H., Luo, Y., & Rabczuk, T. (2020). Hip Fracture Risk Assessment Based on Different Failure Criteria Using QCT-Based Finite Element Modeling. CMC-COMPUTERS MATERIALS & CONTINUA, 63(2), 567-591.

9. Cheng, Y., Zhou, S., Wang, Y., Guo, C., Bai, J., & Tamura, S. (2013). Automatic segmentation technique for acetabulum and femoral head in CT images. Pattern

Recognition, 46(11), 2969-2984.

10. Kim, J. J., Nam, J., & Jang, I. G. (2018). Fully automated segmentation of a hip joint using the patient-specific optimal thresholding and watershed algorithm. Computer methods and programs in biomedicine, 154, 161-171.

11. Almeida, D. F., Ruben, R. B., Folgado, J., Fernandes, P. R., Audenaert, E., Verhegghe, B., & De Beule, M. (2016). Fully automatic segmentation of femurs with medullary canal definition in high and in low resolution CT scans. Medical engineering & physics, 38(12), 1474-1480.

12. Pinheiro, M., & Alves, J. L. (2015). A new level-set-based protocol for accurate bone segmentation from CT imaging. IEEE Access, 3, 1894-1906.

13. Perone, C. S., & Cohen-Adad, J. (2019). Promises and limitations of deep learning for medical image segmentation. Journal of Medical Artificial Intelligence, 2.

14. Hesamian, M. H., Jia, W., He, X., & Kennedy, P. (2019). Deep learning techniques for medical image segmentation: Achievements and challenges. Journal of digital imaging, 32(4), 582-596.

15. Zhu, L., Han, J., Guo, R., Chai W,. Tang S. (2020). An Automatic Classification of the Early Osteonecrosis of Femoral Head with Deep Learning. Current Medical Imaging, 16, 1-9.

16. Chen, F., Liu, J., Zhao, Z., Zhu, M., & Liao, H. (2017). Three-Dimensional Feature-Enhanced Network for Automatic Femur Segmentation. IEEE journal of biomedical and health informatics, 23(1), 243-252.

17. Wang, L., Museyko, O., Su, Y., Brown, K., Yang, R., Zhang, Y., ... & Engelke, K. (2019). QCT of the femur: Comparison between QCTPro CTXA and MIAF Femur. Bone, 120, 262-270.

18. Shorten, C., & Khoshgoftaar, T. M. (2019). A survey on image data augmentation for deep learning. Journal of Big Data, 6(1), 60.

19. Milletari, F., Navab, N., & Ahmadi, S. A. (2016, October). V-net: Fully convolutional neural networks for volumetric medical image segmentation. In 2016 fourth international conference on 3D vision (3DV) (pp. 565-571). IEEE.

20. Deniz, C. M., Xiang, S., Hallyburton, R. S., Welbeck, A., Babb, J. S., Honig, S., ... & Chang, G. (2018). Segmentation of the proximal femur from MR images using deep convolutional neural networks. Scientific reports, 8(1), 1-14.

21. Kingma, D. P., & Ba, J. (2014). Adam: A method for stochastic optimization. arXiv preprint arXiv:1412.6980.

22. Taha, A. A., & Hanbury, A. (2015). Metrics for evaluating 3D medical image segmentation: analysis, selection, and tool. BMC medical imaging, 15(1), 29.

23. Ronneberger, O. , Fischer, P. , & Brox, T. . (2015). U-net: convolutional networks for biomedical image segmentation.

24. Knowles, N. K., Reeves, J. M., & Ferreira, L. M. (2016). Quantitative computed tomography (QCT) derived bone mineral density (BMD) in finite element studies: a review of the literature. Journal of experimental orthopaedics, 3(1), 36.